\documentclass[a4paper,10pt,twoside]{article}
 \usepackage{geometry}
 \usepackage{mathptmx}
\usepackage[disable]{todonotes}
\usepackage{xspace}
\usepackage[utf8]{inputenc}
\usepackage{amsfonts}
\usepackage{enumitem}

\newtheorem{definition}{Definition}
\newtheorem{infdef}{Informal Definition}
\newtheorem{theorem}{Theorem}
\newtheorem{assum}{Assumption}
\newtheorem{example}{Example}

\let\tempone\itemize
\let\temptwo\enditemize
\renewenvironment{itemize}{\tempone\addtolength{\itemsep}{-0.1\baselineskip}}{\temptwo}
\let\tempone\enumerate
\let\temptwo\endenumerate
\renewenvironment{enumerate}{\tempone\addtolength{\itemsep}{-0.1\baselineskip}}{\temptwo}

 \DeclareMathAlphabet\mathcal{OMS}{cmsy}{m}{n}
 \newcommand{\powerset}{\raisebox{.15\baselineskip}{\ensuremath{\wp}}}

 \newcommand{\sym}[0]{\textbf{Sym}\xspace}
 \newcommand{\sen}[0]{\textbf{Sen}\xspace}
  \newcommand{\logic}[0]{\ensuremath{{\mathit{L}}}\xspace}
 
 \newcommand{\intint}[0]{\ensuremath{\mathfrak{I}}\xspace}
 \newcommand{\prop}[0]{\ensuremath{\mathbb{P}} \xspace}
 
 \newcommand{\signature}[0]{\textbf{sign}\xspace}
 \newcommand{\anno}[0]{\textbf{Anno}\xspace}
  \newcommand{\annosmall}[0]{{\tiny{\textbf{Anno}}}}
 \newcommand{\voc}[0]{\textbf{V}\xspace}
 \newcommand{\theo}[0]{\ensuremath{\Gamma}\xspace}
 \newcommand{\atheo}[0]{\ensuremath{\Lambda}\xspace}
 \newcommand{\domain}[0]{\ensuremath{\mathbb{D}}\xspace}
 \newcommand{\val}[0]{\ensuremath{\textit{v}}\xspace}
 \newcommand{\true}[0]{\ensuremath{\textit{true}}\xspace}
 \newcommand{\false}[0]{\ensuremath{\textit{false}}\xspace}
 \newcommand{\ont}[0]{\ensuremath{{\mathbf{o}}}\xspace}
 \newcommand{\sit}[0]{\ensuremath{\mathbb{S}} \xspace}
 \newcommand{\conc}[0]{\ensuremath{\mathbb{C}}\xspace}
 \newcommand{\bpc}[0]{\ensuremath{\overline{\conc}}\xspace}
 \newcommand{\loss}[1]{}

\def\hb{\hbox to 10.7 cm{}}

\begin{document}

 \title{What is an Ontology?}
 \author{Fabian Neuhaus}

 \date{}
 \clearpage\maketitle
 
\begin{abstract}
	In the knowledge engineering community ``ontology'' is usually defined in the tradition of Gruber
	as an ``explicit specification of a conceptualization''. Several variations of this definition
	exist.
	In the paper we argue that (with one notable exception) these definitions are of no explanatory value, because they violate one of the basic rules for good definitions:
	The defining statement (the definiens) should be clearer than the term that is defined (the
	definiendum).
	In the paper we propose a different definition of ``ontology'' and discuss how it helps to explain various phenomena:  
	 the ability of ontologies to change, the role of the choice of vocabulary, the significance of annotations, the possibility of collaborative ontology development, and the relationship between ontological conceptualism and ontological realism.
\end{abstract}

\section{Introduction}\label{sec:intro}

In 1992 Tom Gruber proposed the following definition  ``An ontology is a specification of a conceptualization" \cite{gruber1992}. Several variants exist that usually add adjectives further describing the specification (e.g., ``formal", ``explicit'') or the conceptualization (e.g., ``shared") (see discussion of related work in Section~\ref{sec:related}).
These definitions are not helpful  because they violate one of the basic rules for good definitions: the defining statement (the definiens) should be clearer than the term that is  defined  (the definiendum).
As long as ``conceptualization'' is murkier than ``ontology'', any attempt of defining ``ontology'' as a kind of ``specification of a conceptualization'' is an intellectual placebo:
it makes us feel like it provides a better grasp of the nature of ontologies, but there is no intellectual progress, because it lacks explanatory value (see  Section~\ref{sec:barren} for details).

 Given the difficulties in defining ``ontology'' one may come to the conclusion that a proper definition  is not really needed. This seems to be the sentiment of Gruber in \cite{gruber2009}, who writes after discussing an objection to his original definition:
 \begin{quote}
 	Taking a more pragmatic view, one can say that ontology is a tool and product of engineering and thereby defined by its use.
 \end{quote}
However, it would be rather embarrassing for applied ontologists to give up that easily. After all, we are supposed to be the experts on defining terms. Thus,  we should be able to provide a satisfactory definition of our own subject. The purpose of this paper is to propose a novel definition of ``ontology'' which may be stated informally as:

\begin{infdef}\label{infdef:ontology}
An ontology of a given domain of interest  is a document that provides
	\vspace{-1em}\begin{enumerate}
		\item a  vocabulary for describing the domain of interest,
		\item annotations that documents 
		 the vocabulary, and
		\item a logical theory (consisting of axioms and definitions) for the vocabulary,
\end{enumerate}
	\vspace{-1em} in a way that these three elements together enable a competent user of the ontology to ascertain its intended interpretation.		
\end{infdef}

Before this proposal is presented in detail  and is formalized in Section~\ref{sec:what}, we first discuss some of the phenomena, which should be illuminated by a definition of ``ontology'' (see Section~\ref{sec:requirements}). 
 Central to our account of ontologies is the role of annotations, which
 often not only grounds the `meaning' of the vocabulary, but frequently provides much more information than is captured by the logical theory.
One major benefit of our account is that it explains the relationship between versions of an ontology in different languages (e.g., BFO in FOL and OWL).  It further enables us to clearly analyze certain errors in ontologies, e.g., a {mismatch} between the annotations and the logical theory of an ontology.
The proposed definition has the additional benefit of being compatible with both conceptualism and a realistic reading of ontologies.
Indeed, one can prove that given a correct conceptualization  of a domain, any correct ontology of the conceptualization, is a correct ontology of the domain (Section~\ref{ssec:conceptalismvsrealism}).

\vspace{-1em}
\section{Barren and Fertile Definitions} \label{sec:barren}
We are free to define terms in any way we want, but some definitions are better than others.
To illustrate the point lets consider two different definitions of ``inch''.
\begin{example} { \ }	\label{ex:barley}

	\vspace{-1em}\begin{enumerate}
	\item[E.\ref{ex:barley}a] An inch is the length of three barley corns.\label{ex:b}
	\item[E.\ref{ex:barley}b] An inch is the length of the path travelled by light in a vacuum within
	$\frac{1}{11802852677}$ seconds.
\end{enumerate}

\end{example}
Example E\ref{ex:barley}.a is based on a statute by  Edward II of England in the 14th
century. Example E\ref{ex:barley}.b approximates the contemporary definition,
according to which an inch is 2.54 cm and a meter is defined in terms of the speed of
light. Which of these definitions is better? The answer seems obvious, because
 E\ref{ex:barley}.b provides a better  measurement resolution, since E\ref{ex:barley}.a does not allow measuring length differences below the size of a barley corn. In
 addition,  E\ref{ex:barley}.b enables better measurement precision, since it
avoids measurement uncertainty due to variation in barley
corn size.

However, imagine some  visionary natural philosopher would have proposed to Edward~II to use E\ref{ex:barley}.b  instead of E\ref{ex:barley}.a.
Since  Aristotle discussed the existence of a vacuum
 and the question of the speed of
light\footnote{\cite{aristotleComplete} Physics IV.213a11-217b28,  On the Soul II.418b21-418b26.}, these concepts would be familiar to scholars of the time.  E.g., in the 13th century Witelo considered the possibility that light is faster
when  traveling through a vacuum than when traveling through a thicker medium \cite{marchall1981}.
Thus, assuming
some Arab mathematician would have helped
with the Arabic numerals and fractions, E\ref{ex:barley}.b would not have been
outside the conceptual realm of contemporaries of Edward~II. But should Edward~II have chosen E\ref{ex:barley}.b  instead of E\ref{ex:barley}.a?

Within an agricultural medieval society nobody possessed precise clocks, let alone
the technology to determine how far light travels in a given fraction of a second. Thus, the
contemporaries of King Edward~II would not have been able to apply E\ref{ex:barley}.b  
and use it to determine the length of an object.
Therefore, if  Edward~II would have embraced E\ref{ex:barley}.b, 
this decision would have had no impact on the actual use of the term ``inch''. Since the inch was already established as a measurement unit,  people would have continued to use ``inch'' in
the way they always understood it, while paying lip-service to the definition. Hence,
the royal definition would  not have contributed to a standardization of length measurements. In
contrast,  E\ref{ex:barley}.a provided  a definition of ``inch'' that was easily accessible and usable  for people in an agricultural society. 

This example illustrates that the {fertility} of a definition depends on the
context of its use. \emph{Fertility} denotes the degree to which people who embrace a
definition are able to utilize it, e.g., to classify some phenomenon or solve a problem
or to make some other tangible 
intellectual process. In the context of the 21st century
E\ref{ex:barley}.b is more fertile than E\ref{ex:barley}.a, because  it enables better
measurements of lengths and connects it systematically to other units of measurement in the International System of Units. However, in the context of a medieval society
E\ref{ex:barley}.b  would serve no useful function at all;
it would be \emph{barren}. This is the case, because King Edward's
contemporaries already had  a better understanding of ``inch'' (the definiendum) than  of the definiens of  E\ref{ex:barley}.b. 
\loss{\footnote{Barrenness of a definition  is closely related to the Fallacy of Obscurity (\cite{copy}, p. 156). The difference is that the definiens does not contain metaphorical language,  but less informative language.}}

Most definitions of ``ontology'' are as barren as E\ref{ex:barley}.b would have been  for King Edward~II, and for the same reasons.
It is quite easy to introduce the term ``ontology'' to a novice to the  field by illustrating it with
examples and use cases. Such an introduction leads to an operational understanding of the term
``ontology'', which allows the novice to recognize typical examples of ontologies and to recognize
typical situations where ontologies may be used to solve a problem. (Analogously to the members of King Edward's II court, who
had an operational understanding of ``inch''.)
Thus, any definition of ``ontology'' may only be fertile, if our understanding of the definiens exceeds this kind of operational
level of understanding of ``ontology''. However, the terms ``concepts'' and  ``conceptualization'' are notoriously hard to explain. Indeed, the nature of concepts is an open question in  philosophy \cite{sep-concepts, nimtz2010}. This is reflected by the diverse explanations of ``conceptualization'' in the applied ontology literature, which include:
 \begin{itemize}
	\item ``the objects, concepts, and other entities that are assumed to exist in some area of interest and the relationships that
	 hold among them\loss{ (Genesereth \& Nilsson, 1987)}'' \cite{gruber1995}
 	\item ``[...] abstract, simplified view of the world that we wish to represent for some purpose'' \cite{gruber1993}
	\item ``an intensional semantic structure which encodes the implicit rules constraining the structure of a piece of reality'' \cite{guarino1994}
	\item ``a structured interpretation of a part of the world that people use to think and communicate about the world'' \cite{borst}
	\item ``an abstract model of some phenomenon in the world by having identified the relevant concepts of that phenomenon'' \cite{studer1998}
 \end{itemize}
Thus, in the context of the prevalent fuzzy understanding of the term ``conceptualization'',
any definition of ``ontology'', which is based on it, is as barren as a definition of
``inch'' based on the speed of light in the 14th century.
Such a definition may serve as a scholarly ornament, but it will not  fulfill an explanatory function.\footnote{A notable attempt to address this issue and  clarify ``conceptualization'' with the help of a formal theory was initially  proposed by Guarino in \cite{guarino1998} and further developed in \cite{guarino2009}. See Section~\ref{sec:related}.}

\section{Towards a Fertile Definition of ``Ontology''}\label{sec:requirements}
This section enumerates a number of various phenomena that a fertile definition
of ``ontology'' should be able to illuminate.

\subsection{Ontologies, Versions, and Sets of Formulas}
According to Guarino et al. ontologies are logical theories -- i.e., sets of formulas -- that meet certain conditions \cite{guarino2009}.
Undoubtedly, ontologies involve logical theories, but there are reasons not to simply identify an ontology  with a set of formulas.
First of all, ontologies change over time and, thus, different versions of
 the same ontology may be associated with different sets of formulas
 (e.g., because additional axioms have been added).  Further, the same ontology may be
 maintained in different languages (e.g., OWL and FOL). One cannot even identify an
 ontology  version  with a set of formulas, because different ontology versions may
 differ with respect to their annotations (e.g., if a natural language definition of a term
 is added). Any explanation of ``ontology'' needs to address the relationships between ontologies, their versions,
 and logical theories.\footnote{This fact was pointed out to me by Barry Smith  in a conversation.}

\subsection{The Importance of the Vocabulary and Annotations } \label{ssec:importance}

From a logical point of view there is no significant difference between the logical
theories E.\ref{ex:voc}a-E.\ref{ex:voc}d, because for any OWL
interpretation  that satisfies one of them, there are isomorphic interpretations that satisfy the others.

\begin{example} { \ }	\label{ex:voc}
	\vspace{-1em}\begin{enumerate}
	\item[E.\ref{ex:voc}a] Class: P  DisjointWith: R
	\item[E.\ref{ex:voc}b] Class: Dog  DisjointWith: GermanShepherd
	\item[E.\ref{ex:voc}c] Class: Biolgy DisjointWith: Chenistry
	\item[E.\ref{ex:voc}d] Class: Biology DisjointWith: Chemistry

\end{enumerate}
\end{example}

From an ontological point of view E.\ref{ex:voc}a-E.\ref{ex:voc}d could not be more different.
E.\ref{ex:voc}a is a logical theory, but it is not representing any knowledge about any domain.
Thus, E.\ref{ex:voc}a is not an ontology.
In contrast, in the absence of additional information, it is reasonable to assume that E.\ref{ex:voc}b is an ontology that is about dogs, and the ontologies E.\ref{ex:voc}c and E.\ref{ex:voc}d are  about the
relationship of two sciences. The only difference is that E.\ref{ex:voc}c contains spelling errors, while E.\ref{ex:voc}d does not.

Note that, from a logical point of view, E.\ref{ex:voc}d stands in exactly the same relationship to E.\ref{ex:voc}c as to E.\ref{ex:voc}b. Nevertheless, it seems obvious to us that  E.\ref{ex:voc}d and  E.\ref{ex:voc}b are about different domains, while
 we assume -- at least in the absence of additional information -- that  E.\ref{ex:voc}d and  E.\ref{ex:voc}c are about the same domain.  Further, we know that axiom  E.\ref{ex:voc}b  is false, while the content of  E.\ref{ex:voc}c is true but the spelling of
the vocabulary needs to be fixed.

We are able to make these distinctions not because of any logical properties of the different logical theories
 E.\ref{ex:voc}a-E.\ref{ex:voc}d, but because the choice of the vocabulary is an essential component of an ontology. By using terms
 from a given natural language we establish the \emph{intended interpretation} of  the ontology and, indirectly, the domain
 that the ontology is about. If we fail to establish such an intended interpretation as in the case of E.\ref{ex:voc}a, there is no ontology.

 \medskip

Annotations provide another tool for establishing intended interpretations.
We use the term ``annotation'' in the broadest possible sense that includes any kind of material that is included in an ontology with the intent to document its vocabulary or its logical axioms. This may include  natural language definitions,
explanations, comments, examples, references, links, diagrams, pictures, audio files or video files.

Annotations are important, because the use of individual terms as part of a vocabulary may leave room for different interpretations and
ambiguity. E.g., the logical content of E.\ref{ex:annotation}a does not differ from example E.\ref{ex:voc}d. However, the annotations
 clarify  that the term ``biology'' is not intended to refer to biology, but to classes about biology. Note that the annotations
in E.\ref{ex:annotation}a contain information that is not captured in the axiom:
the first annotation provides a label for the class,
the second attributes the axiom to its creator,  the third includes a warning.

These examples illustrate why, typically, the axioms of an ontology reflect only a fraction of the
information that is provided by the annotations. We need to distinguish   {assertive annotations} from the other annotations.
\emph{Assertive annotations} are the kind of annotations that are intended to assert a true
 proposition about the domain of the ontology. E.g., adding a new label to a class does not involve
  an assertion. The inclusion of metadata (e.g., authorship) is an assertion, but about the ontology and not about the domain of the ontology. Thus, the second annotation of  E.\ref{ex:annotation}a  is not an assertive annotation. The third annotation is a warning. This particular warning is a speech act that involves an assertion about the domain, namely, that for some material parental approval is required. Hence, the last two annotations of  E.\ref{ex:annotation}a  are assertive.

The logical theory of an ontology may only represent the content
of assertive annotations. The third annotation of  E.\ref{ex:annotation}a illustrates that often some assertive annotations are not axiomatized.
This may be the case for various reasons. E.g., because the
formalization would require an excessive extension of the vocabulary or the formal language is not expressive enough. In any case, a
reductionistic view on ontologies that treats them as mere logical theories is prone to ignore much of the information they contain.

\begin{example} { \ }	\label{ex:annotation}

	\vspace{-1em}\begin{enumerate}
	\item[E.\ref{ex:annotation}a] Class: Biology   DisjointWith: Chemistry \\
								  Annotations:
								  rdfs:label ``Biology classes for middle school'', creator Fabian, \\
								  rdfs:comment ``Warning: involves material that requires parental approval'', \\  rdfs:comment ``No biology classes are chemistry classes''

	\item[E.\ref{ex:annotation}b] Class: P  Annotations: rdfs:label ``Dog'' DisjointWith: R \\
						   Class: R  Annotations: rdfs:label ``German Shepherd''

	\item[E.\ref{ex:annotation}c] Class: Dog  DisjointWith: GermanShepherd  \\
  Annotations:   rdfs:comment ``German Shepherd is a popular breed of dog''
\end{enumerate}
\end{example}
Example E.\ref{ex:annotation}b illustrates that annotations are sufficient to establish an intended interpretation, even if the
vocabulary consists of arbitrary symbols. Note that the logical theory of E.\ref{ex:annotation}b  is identical to E.\ref{ex:voc}a, but with the help of the annotations the theory is turned into an ontology.
Indeed, one could argue that E.\ref{ex:annotation}a and E.\ref{ex:voc}b are \emph{ontologically equivalent},
although their logical theories are clearly not logically equivalent.

Example E.\ref{ex:annotation}c shows that there may be a \emph{mismatch} between the logical theory and the annotations of an ontology. It is not possible to detect this kind of problem with the help of an OWL reasoner, since the logical theory is logically consistent.
The problem arises because the annotation contradicts the annotation.

The previous examples show that the choice of the vocabulary and the annotations play an important role, since in tandem with the logical theory they  allow  the user of the ontology to 
establish the intended interpretation of an ontology. By the latter we mean that the user is able to identify the domain of the ontology and  to understand the propositions that are asserted about the domain by the annotations 
and the logical theory. 
As it is the case for any complex texts, 
understanding an ontology is typically a hermeneutic process, which leads from a very preliminary understanding of the ontology to a situation where one understands the big picture as well as all the details. The success of such a process depends 
not only on the quality of the ontology, but also on the knowledge of the user. If, for example, the vocabulary and the annotations are in English, they will be of limited use to a user who does not speak English. %
\todo{When in the process is the threshold for establishing intended interpretation reached?}
\todo{intended interpretation does not entail knowledge whether true or false}
A good definition of ``ontology'' should reflect that the choice of the vocabulary and the annotations play an important role. Further, it should enable an explanation of ontological equivalence and mismatch between axioms and annotations.

\subsection{Sharing an Ontology}\label{ssec:sharing}
Starting with Gruber  \cite{gruber1993}, the role of ontologies for sharing knowledge  featured prominently in discussions on the
nature of ontologies.
Two important benefits of an ontology are that it may be used as a knowledge resource
for different applications, and that the ontology
 enables different people (and systems) to exchange information.
For this reason an ontology is often defined as a specification that specifies a ``shared
conceptualization'' (see Section~\ref{sec:related}). This is problematic for two reasons.

Firstly, a philosophical challenge arises if one assumes that a conceptualization is a kind of mental representation
of a domain that is built based on experiences.
As \cite{guarino2009} points out,
under this assumption it is hard to see how conceptualizations may be shared in the literal sense of the word.
Because mental entities
 are necessarily private in the sense that they are not accessible to anyone except the person who experience them.
\loss{For example, if I have a tooth pain, nobody else may experience the same tooth ache. I could not share the tooth ache experience any
more than I could sell it to or borrow it from somebody else. Of course, I could offer somebody else the experience of listening to me
complaining about my tooth ache, but that is an altogether different experience. The same
argument applies to other mental entities (e.g., emotions, perceptions, sensations). It seems dubious why conceptualizations should behave differently.}

Secondly, the claim that ontologies are based on a shared conceptualization contradicts   the \emph{intellectual division of labor} that is typical for the development of large ontologies.
Imagine that a manufacturer of cell phones develops an ontology in order to represent knowledge about its products across their whole
life-cycles. Thus, the ontology needs to represent, for any phone, its design (form, size, haptic features), its electronic
parts and their functions, the topology of the components, a plan of how the phone is assembled, and information about repair and recycling. Obviously, every of these aspects is developed by specialists. Let's assume that the ontology development team consists of experts from the various areas.
Typically, the designers of the phone have  little knowledge about the assembly plan,
while the people who are responsible for recycling typically won't know details about the functional specifications of the parts.
Hence, the conceptualization of the cell phone domain by any member of the ontology development team is very detailed in the area that the person specializes in, and shallow in
other areas. 
Thus, the conceptualizations of the domain of the ontology varies significantly
across its developers.  Therefore, there is no single conceptualization of the domain that is shared by everybody.
Indeed, since the ontology will contain more information than any individual team member possesses, the ontology
specifies no single person's conceptualization.

While the developers of an ontology typically will possess different conceptualizations of the domain, there must be some harmonization, otherwise the development of the ontology will fail. How do we explain this?

\subsection{Conceptualism vs. Realism}
Ontological realists object against any definition of ``ontology'' as a specification of a conceptualization because -- according to them -- ontologists are not about conceptualizations but
 about reality  \cite{smith2004}. This is not the place to recapitulate the arguments in the debate between ontological realists and conceptualists. However, since either position is reasonable, it
 seems desirable to define ``ontology'' in a way that is compatible with either position and, 
ideally, clarifies the exact difference between them.

\vspace{-1em}
\section{What is an Ontology?}\label{sec:what}
\subsection{Ontologies are Documents} \label{ssec:documents}
An ontology is a kind of document, like a book or a schedule of a conference. Because ontologies are documents, the same kind of systematic ambiguities that apply to other documents apply to ontologies. E.g., in the sentences ``The conference schedule was fleshed out in our last planning meeting'' and ``I downloaded the conference schedule'' and ``You will find a conference schedule among your meeting materials'', the expression ``conference schedule'' denotes three different entities. In the first sentence the  conference schedule  is an entity that may evolve over time, the second sentence is about a particular snapshot or version of the first entity, and the third is about particular printouts (or tokens)  of the second entity. The same ambiguities
apply to ontologies, e.g., consider ``The Gene Ontology changed significantly since 2003. I downloaded it yesterday, it is in my home directory on my laptop.''

These kind of systematic ambiguities are benign and we are so used to them that  we usually do not notice. But given that the goal of this paper is a clear definition of the term ``ontology'', it
 is worth to avoid ambiguity. Thus, for the purpose of this paper we distinguish between documents and document versions\footnote{Both are distinguished from document tokens (e.g., particular
   files on a hard drive or particular printouts). Document tokens are not relevant for the purpose of this paper.}. This distinction follows the United Nations System Document Ontology (UNDO)
   \cite{peroni2017undo}, according to which a document is realized by its versions. A particular version of a document may be derived from another version in different ways. The UNDO distinguishes between revisions, translations, and transformations.

The distinctions for documents in general also apply to ontologies. An \emph{ontology version} is a realization of an ontology.
An ontology version may be derived from a previous version in various ways, e.g., by adding new axioms, removing annotations, or correcting spelling mistakes. Further, an ontology
 version may be the result of translating the annotations in a different language (e.g., from English to Spanish) or by translating the
 logical theory from one logic to another (e.g., from OWL to first-order logic) or by changing the serialization of the logical theory
 (e.g., from a serialization in OWL Manchester Syntax to Turtle).
 Thus, ontology versions do not exist in isolation, but form a network. This network can be thought of as a directed acyclic graph where the edges represent the derives-from-relationship. This leads us to Definition~\ref{def:ontology}.

\begin{definition}[Ontology]\label{def:ontology}
An 	ontology of a domain is a document that is realized by a network of ontology versions about the domain. 
\end{definition}
Of course, this definition only pushes our original question one step further and we need to ask ourselves: \emph{What is an ontology version about a domain?}

\subsection{Ontology versions}
As we have seen in Section~\ref{ssec:importance},
an ontology version involves three elements: a vocabulary, annotations, and a logical theory in some logic; and  the vocabulary and the annotations play an important role in establishing the intended interpretation of an
 ontology.
For anyone who attempts to analyze these observations closer, one challenge is that one has to avoid too specific assumptions about the entities involved. The vocabulary most likely consists of
strings, but even that is not necessarily the case.
 As discussed above, annotations come in a wide
variety of forms, some of them are  assertive and some are not.
And since ontologies are written in various logical formalisms, we cannot make any strong assumptions about the logic either. Most importantly, one needs to address the nature of the rather mystical `intended interpretation'.

Let's start with the basics.
By a  \emph{domain of interest} \domain{}  we understand  a topic that one or more people may be interested in. Examples include quantum physics, FC Liverpool, cardinal numbers, and Game of Thrones.

 By \emph{propositions} we understand  the semantic
contents of  assertive sentences. Propositions may be true or false, and they are subjects of beliefs. It seems plausible to assume that the proposition that is asserted by a complex
 sentence like ``John is bald and Fred is tall'' is structured in some way and is connected
 in some interesting
fashion to the proposition that is asserted by ``John is bald''. However, within
the context of this paper we abstract from the internal structure of propositions and do not worry about how they relate to each other.
The only exception is that we assume that any set of propositions is either \emph{consistent} or \emph{inconsistent}.
Two sets of propositions $\prop_1,\prop_2$ are \emph{weakly equivalent} iff, for any set of propositions $\prop_3$,  $\prop_1 \cup \prop_3$  is inconsistent iff $\prop_1 \cup \prop_3$ is inconsistent.
 For any given domain \domain, $\prop_\domain$ is the set of propositions that is about \domain{}.
 Assume $P^{\dagger} \subseteq \prop_\domain$.  A consistent set of propositions $P \subseteq  \prop_\domain$ is
 a \emph{maximal consistent} subset of $P^{\dagger}$ iff there is no consistent set of
   propositions $P^\prime$ such that $P\subset P^\prime$ and $P^\prime \subseteq P^{\dagger}$.

Since we do not want to consider a particular logic for our purpose, we define a logic as an entailment system.\footnote{As \cite{avron1991simple} points out, in order to include formalisms like linear logic, one would have to use  multisets instead of sets. Def.~\ref{def:logic} is richer
than the corresponding definition in  \cite{avron1991simple}, since the chosen vocabulary is important for ontologies. One other difference is that Def.~\ref{def:logic} allows the set of premises to be infinite.}
\begin{definition}[Logic] \label{def:logic}
A logic  $\logic =\langle \sym, \sen , \signature ,  \vdash \rangle$ consists of
\vspace{-1em}
\begin{itemize}
	\item the set of symbols $\sym$;
	\item the set of sentences \sen over  $\sym$;
	\item  the function $\signature  :   \powerset(\sen) \rightarrow \powerset(\sym) $ that maps a set of sentences to its signature  (i.e., the set of symbols that occur in it), such that \begin{itemize}
	\vspace{-0.5em}
		\item $\signature(\emptyset) = \emptyset$ and  $\signature(\sen) = \sym$;
		\item if $\Gamma \subseteq \Delta \subseteq \sen$, then $\signature(\Gamma) \subseteq  \signature(\Delta)$;
	\end{itemize}
	\vspace{-0.3em}
	\item an entailment relationship $\vdash \  \subseteq \ \powerset{}(\sen) \times \sen \ $.
\end{itemize}
\end{definition}


Based on our definition of a logic, we may introduce the notion of an annotated logical theory.
For any logic $\logic=\langle \sym, \sen , \signature, \vdash \rangle$,
\theo is a theory in \logic iff $\theo \subseteq \sen$;   \theo is a
   theory over \voc in \logic iff $\voc \subseteq \sym$ and \theo is a theory in \logic and $\signature(\theo)\subseteq \voc$.

\begin{definition}[Annotated Logical Theory]

	An annotated logical theory $\langle \voc, \theo,\allowbreak \anno, \logic \rangle$ consists of
\vspace{-1em}
	\begin{itemize}
		\item   a non-empty set of symbols  $\voc$ (the vocabulary);
		\item a theory $\theo$ over $\voc$ in \logic;
		\item a set $\anno$  of annotations, which  is partitioned into the set of assertive annotations $\anno^a$ and non-assertive annotations $\anno^n$;
				\item a logic \logic{}. 
		\end{itemize}

\end{definition}

 Given a set of assertive statements about a domain, their intended interpretation may be considered as a mapping from these syntactic entities to propositions about the domain. 
 In the case
 of the ontology these assertive statements come in two flavors: the assertive annotations and
 the axioms of the ontology.

 \begin{definition}
 Let $\atheo = \langle \voc, \theo, \anno, \logic \rangle$ be an annotated logical theory and let \domain be a domain. 	An intended interpretation $\intint$ from \atheo{} into \domain is a mapping such that: (1) for any annotation $a \in \anno^a$, $\intint(a)\in \prop_\domain$; and (2)
for any sentence $\phi \in  \theo $, $\intint(\phi)\in \prop_\domain$.
 \end{definition}

As we have discussed in  Section \ref{ssec:documents},  an ontology version is a kind of document
version. Now, we can be more specific: the authors of an ontology version specify an annotated theory in such a way that the users are able to grasp its intended interpretation 
(see Definition~\ref{def:ontvers}). This usually involves the use of an established terminology for the domain either as the  vocabulary or as labels for the vocabulary. 
Note that the ontology version does not necessarily need to contain the complete vocabulary,  axioms, and annotations of the annotated logical theory, since the ontology version may contain importations.

\begin{definition}[Ontology Version] \label{def:ontvers}
Let \domain be a domain. \\
\ont is an \emph{ontology version} about \domain iff \ont is a document version that specifies exactly one annotated logical theory $\atheo = \langle \voc, \theo, \anno, \logic \rangle$ such that the choice
 of the the vocabulary \voc together with the annotations in \anno
 and the axioms in $\Gamma$
enable a competent user of the ontology to  ascertain the intended interpretation
function \intint from \atheo into \domain.

\noindent Let $\prop^\ont$ be the set of propositions asserted by \ont, where
$\prop^\ont  =  \prop^\ont_\annosmall \cup \prop^\ont_\theo$ and  
$\prop^\ont_\annosmall = \{ \ p \in \prop_\domain
   \ |  \ p = \intint(a), a \in \anno^a \} $ and   $\prop^\ont_\theo = \{ \ p \in \prop_\domain \ |
	\ p = \intint(\phi) \textrm{ and } \theo
    \vdash \phi \}$.

\end{definition}
In  Definition \ref{def:ontvers}   $\prop^\ont_\annosmall$ is  the set of propositions that are the intended
 interpretations of the assertive annotations of the  ontology \ont{}. Analogously,  $\prop^\ont_\theo$ is the set of
 propositions that are intended interpretations of formulas that are entailed by $\Gamma$.
   Definition \ref{def:ontvers} allows us to define various notions. Let
  $\ont_1,\ont_2 $ be ontologies about a domain \domain.   $\ont_1,\ont_2 $ are \emph{strongly equivalent} iff $\prop^{\ont_1} = \prop^{\ont_2}$.
$\ont_1,\ont_2 $ are \emph{weakly equivalent} iff  $\prop^{\ont_1}$ and $\prop^{\ont_2}$ are weakly equivalent.
$\ont_1$ is \emph{stronger} than  $\ont_2$ iff
there is some $\prop^\prime \subseteq \prop_\domain$ such that  $\prop^{\ont_1}$ and $\prop^{\ont_2} \cup \prop^\prime$ are weakly equivalent.
 \ont contains a \emph{mismatch between the logical theory and the annotations}  iff  $\prop^\ont_\annosmall$ and  $\prop^\ont_\theo$  each are consistent, but $\prop^\ont_\annosmall \cup \prop^\ont_\theo$  is inconsistent.

These definitions   allows us to explain  the examples E.\ref{ex:voc} and E.\ref{ex:annotation}.  E.\ref{ex:voc}a is not an ontology version, since it does fail to establish an intended interpretation. The intended interpretations of  E.\ref{ex:voc}b and E.\ref{ex:voc}d map their axioms to different propositions.
 In contrast, E.\ref{ex:voc}c and E.\ref{ex:voc}d  are mapped to the same proposition, and, thus, they are   {strongly ontologically equivalent}.  In contrast to  E.\ref{ex:voc}a, the  annotations in E.\ref{ex:annotation}b specify the intended interpretation. E.\ref{ex:annotation}b  and  E.\ref{ex:voc}b  are ontologically strongly equivalent.
E.\ref{ex:annotation}c contains a mismatch between its annotations and its logical theory.

Definition~\ref{def:ontvers} does not assume that the whole document is written in some formal
language. We do not include this requirement, since there are well-known publications of ontologies that do not meet it. E.g., in \cite{masolo2003ontology} DOLCE is specified in a PDF document that contains free
text and axioms. Of course, the axioms are written in a formal language, but the overall
document is not.

According to Definition \ref{def:ontvers} an ontology is basically an annotated logical theory which is interpreted as propositions about a domain. These propositions do not need to be true or correspond to a conceptualization. Thus, even an ontology that contains only false axioms is an ontology. How ontologies relate to conceptualizations or reality will be discussed next.

\subsection{Ontologies from a Realists' Perspective}
According to ontological realism, an ontology is about a particular slice of reality (on a given level of granularity viewed from a
particular perspective) \cite{smith2004}. Let's call a domain of interest that meets that description a  {real domain}. The anatomy of zebrafish and the political structure of Iran are examples for real domains. In contrast, the anatomy of unicorns or the political
 structure of  the Seven Kingdoms in the Game of Thrones world are not real domains. Thus,  from a realist's perspective these topics are no suitable domains of interest for an
ontology.
This is a major difference to the conceptualist's position, whose domain of interest may involve fictional entities.

One way to characterize a realistic domain is by the different states of affair that
may come about in this domain. A state of affair may be formally represented by maximal consistent
sets of propositions about the domain. Thus, for the purpose of this paper (admittedly, a strong
simplification) we assume the following:
%
	A \emph{real domain} \domain is a set of maximal consistent subsets of $\prop_\domain$.
Based on this definition, we may define when an ontology of a real domain is correct:
\begin{definition}[Correct Ontology] \label{def:correctReal}
	A document  \ont is a correct ontology version of a real  domain \domain  iff  \ont is an ontology version about \domain and for any  state of affair $s \in \domain$:  $\prop^\ont \subseteq s$.
\end{definition}

\subsection{Ontologies from a Conceptualist's Perspective}\label{ssec:conceptper}
According to ontological conceptualism, ontologies specify conceptualizations (of some domain).
As discussed in  Section~\ref{sec:barren}, the nature of conceptualizations is
unclear.  However, we will make the following assumption about conceptualizations: if an agent internalizes a
conceptualization \conc of a given domain \domain{}, then the conceptualization enables the agent when presented a given state of
affairs $s$ to evaluate for a given proposition $p\in \prop_\domain$ whether (a) $p$ is true in $s$ according to \conc or (b) $p$ is false in $s$
according to \conc or (c) whether the truth-value of $p$  in $s$ is not determined by \conc.
This assumption allows us to avoid the thorny question of the nature of conceptualizations. Instead we treat
them as black boxes that are characterized by the ability of an agent who  internalizes the conceptualization to answer questions (or
show some other behavior that allows us to infer how the agent evaluates propositions).
Since this is a kind of behavioristic turn, we call this characterization the  \emph{behavioral profile of a conceptualization} (BPC) of a given domain.
Such a BPC consists of the set of states of affair in the domain that the conceptualization is about
and provides for each state of affair a partial mapping from propositions to \true or \false. The mapping is partial, since we allow incomplete conceptualizations of a domain.

\vspace{-0.5em}
\begin{definition}[Behavioral Profile of a Conceptualization (BPC)]
	A BPC of a domain \domain is an ordered pair $\langle \sit, \val \rangle $ where
\vspace{-1em}
	\begin{enumerate}
		\item the set of states of affair \sit{} is non-empty;
		\item the valuation function \val assigns a partial function $\val_s :   \prop_\domain \rightarrow \{ \true , \false \}$, to each state of affair  $s\in \sit$. 
	\end{enumerate}
\end{definition}
\begin{assum}
	For any conceptualization \conc of a domain \domain there exists exactly one BPC \bpc of \domain that characterizes the ability of an agent that internalizes \conc to evaluate propositions about \domain.
\end{assum}
Based on this assumption, we could define a  ontology version to be a correct specification of a given conceptualization if
the  asserted propositions  are true under any state of affair that is part of the conceptualization. However, as we have discussed in Section
  \ref{sec:requirements}, the development of ontologies typically involves a number of people with varying conceptualizations. For this
  reason, we generalize the notion as follows:  an ontology version is a correct specification of a set of conceptualizations, if it is consistent with each individual conceptualization and every proposition that is asserted by the ontology is true according to at least one conceptualization (see Def.~\ref{def:correctSpec}). An ontology version may contain bugs. Thus, not every specification by an ontology version is correct. Nevertheless, the ontology version is still a specification of the conceptualizations (see Def.~\ref{def:specConc}).\todo{clumsy, rewrite}

\vspace{-0.5em}
\begin{definition}[Correct Specification of  Conceptualizations]\label{def:correctSpec}
	Let  \domain{} be a domain,
	$\conc$ a  conceptualization of  \domain{} such that $\bpc  = \langle \sit, \val \rangle$,
	$\mathfrak{C} = \{\conc_1, \ldots, \conc_n \} $ a set of conceptualizations  of  \domain{}, such that for any $\conc_i \in \mathfrak{C}$
	the corresponding BPC  $\bpc_i  = \langle \sit_i, \val_i \rangle$,
and	 \ont an ontology version about \domain.

\noindent
\ont is consistent with $\conc$ iff  \ont is an ontology version about \domain and for any  state of affair $s \in \sit$:
$\val_s(p) \neq \false$, for any $p \in \prop^\ont$.\\
\ont is  a correct specification of $\mathfrak{C}$ iff  \ont is an ontology version about \domain and

\vspace{-1em}
\begin{itemize}
	\item \ont is consistent with $\conc_i$, for any $\conc_i \in \mathfrak{C}$;
	\item for any $p \in \prop^\ont$, there is a $\conc_i \in \mathfrak{C}$, such that for all $s\in \sit_i$,  $\val_{s}(p) = \true$.
\end{itemize}
\end{definition}
%
%
\vspace{-1.5 em}\begin{definition}[Specification of  Conceptualizations] \label{def:specConc}
An ontology version   $\ont$  specifies the conceptualizations $\mathfrak{C}$ iff the authors of \ont intend \ont to be a correct specification  of $\mathfrak{C}$.
\end{definition}
Definitions \ref{def:correctSpec} and \ref{def:specConc} explain why a division of  intellectual labor does not prevent ontology development,
because it is possible for people to develop an ontology together, even if they do not share the same conceptualization of the domain.
The different ontology developers may even possess conceptualizations of the domain that contradict each other, as long as the ontology
is non-committal concerning the contested propositions.

\subsection{Conceptualism vs. Realism}\label{ssec:conceptalismvsrealism}
While the conceptualist raises the question whether an ontology accurately represents our conceptualizations of a given domain,
the realist is mainly concerned with the question whether an ontology represents reality  accurately. Both questions are quite
reasonable and related. We call a conceptualization of a real domain correct, if in
all situations that are considered by the conceptualization, the valuation function in that situation does not conflict with reality (see Def.~\ref{def:correctConc}).
One can prove that  a correct specification of a correct conceptualization, is a correct ontology of the domain (Theorem~\ref{trm:correctness}).

\begin{definition}[Correct conceptualization]\label{def:correctConc}
Let \domain be a real domain, $\conc$ be a conceptualization of \domain, and  let $\bpc = \langle \sit, \val \rangle$.

\noindent $\conc$  is a correct conceptualization of  \domain iff \
(1) $\sit \subseteq \domain$ and
(2) for any state of affair $s \in \sit $ and any proposition $p \in \prop_\domain$:
\vspace{-1em}	\begin{itemize}
		\item[(a)] if $\val_s(p) = \true$, then $p\in s $;
		\item[(b)] if $\val_s(p) = \false$, then $p\not\in s $.
	\end{itemize}
\end{definition}

\begin{theorem}\label{trm:correctness}
	If 	$\mathfrak{C}$ is a set of correct conceptualizations of the real domain \domain
	and \ont is a correct specification of $\mathfrak{C}$,
	then \ont is a correct ontology version of \domain{}.
\end{theorem}

\vspace{-2em}
\section{Related Work}\label{sec:related}
In the context of information systems
  ``ontology'' was first used in 1967 by Mealy   \cite{mealy1967data}.\footnote{
For the sake of brevity, we limit our discussion of the history of the term ``ontology'' and, in particular, omit its history in philosophy. A more detailed account  can be found in \cite{smith2004InfoSys}.}
However, arguably more significant is \emph{Some Philosophical Problems from the Standpoint of Artificial Intelligence} \cite{mccarthy1969}
from 1969, in which McCarthy and Hayes argue that general intelligent machines require  \emph{metaphysically and epistemologically adequate representations} of the world,
and propose the  situation calculus  as a representation language.  Although the term is not used in \cite{mccarthy1969}, today we would call these representations `ontologies'. The vision presented in  \cite{mccarthy1969} lead to Hayes' work on naive
physics, 
in particular his 1978 paper ``Naive Physics I: Ontology for Liquids'' \cite{hayes1978liquid}, which
seems to be the first computer science paper with ``ontology'' in the title.
\cite{hayes1978liquid} does not contain an explicit definition of ``ontology'', but the first sentence states that the essay contains ``[...] a formalisation of common-sense knowledge [...]''. \loss{This formulation was echoed by early descriptions of Cyc \cite{}.}

Thus, the term ``ontology'' was  already established in computer science, when in 1992 Tom Gruber proposed the
definition ``An ontology is a specification of a conceptualization'' \cite{gruber1992}. Shortly afterward
Gruber added the condition that the specification is explicit
 \cite{gruber1993, gruber1995},  Borst \cite{borst} added the
 conditions that the specification needs to be formal and the
 conceptualization needs to be shared.  Studer
 \cite{studer1998} combined these and defines an ontology as a 
 \emph{formal, explicit specification of a shared  conceptualization}.

All of the modifications of Gruber's initial definition are  dubious.
In \cite{gruber1993} it is quite clear that Gruber
borrows the term ``specification'' from the formal software specification community, which uses formalisms like ``Z'' to
describe software systems. In
\cite{gruber1993} KIF 
is used as specification language.
Thus, any specification in the sense of Gruber is formal.
Further, since it is unclear what an implicit KIF document would look like, the explicity condition seems
redundant as well.\footnote{Unfortunately, in \cite{gruber1993, gruber1995} Gruber does not discuss  why he added the explicity condition.}
 Borst argues for including the shared-conceptualization-condition, because ``[...] the ability to reuse an ontology will be almost nil
  when the conceptualization it specifies is not generally
  accepted.'' \cite{borst}.
The argument seems to be based on the underlying assumption that a shared conceptualization is a precondition for shared use or reuse. As we have argued in Sections~\ref{ssec:sharing}
and \ref{ssec:conceptper}, an ontology may represent knowledge from people whose conceptualization of the domain differs. 

Guarino argues that the term
``specification'' is too strong and suggests to weaken the definition to ``logical theory which gives an explicit, partial account of a conceptualization''  \cite{guarino1995}.
 Smith and Welty argue that Gruber's definition is too wide, since it encompasses a diverse group of artifacts ranging from catalogs, over glossaries to a set of logical constraints. \cite{smithwelty2001}

One notable exception to our criticism of conceptualization-based definitions of ontology is the seminal work by Guarino et al. in \cite{guarino2009}, which builds on \cite{guarino1995, 
guarino1998}.  It presents a definition of ``ontology'' based on a formal  theory of conceptualizations.  In short, \cite{guarino2009} identifies  an ontology with a set of formulas whose models approximate a conceptualization, which is identified with an intensional  first-order  relational structure. 
Since \cite{guarino2009} identifies ontologies with sets of formulas, it does not 
explain the ability of ontologies to change, and fails to address the role of the choice of the 
vocabulary and the annotations. Further, it also assumes that ontologies are based on shared conceptualizations, which fails to account for the division of labor during ontology development. Two other problematic aspects are the lack of explanation of what ``approximates'' is supposed to mean, and the identification of conceptualizations with intensional 
first-order
structures that lead to several technical and philosophical issues. 
A detailed critique of \cite{guarino2009} may be found in \cite{neuhaus2017}.

\section{Conclusion}
In this paper we presented an alternative definition of ``ontology''. It is based on a formal theory that involves two key notions: an \emph{annotated logical theory}  and an \emph{intended interpretation}. 
The definition  does not make any assumptions about the logical language of the ontology or 
the type of annotations that are used. Further, it is not only compatible with both conceptualism 
and realism, but allows showing that a correct specification of correct conceptualizations  is a 
correct ontology (version). The fruitfulness of the definition is further illustrated by an explanation of  the role 
of the  choice of the vocabulary and annotations for establishing an intended interpretation, and an account of how  cooperative ontology development is possible without a shared conceptualization. 

In the future we are planning to study the approach's consequences for ontology evaluation (completeness of axioms, underaxiomatization, lack of documentation). Further, we are planning to instantiate it for particular logics (e.g., FOL), which will enable us to relate the formal semantics of the logic to the intended interpretations. 

\vspace{-1em}
\section*{Acknowledgments}
I am very thankful for helpful feedback from 
Martin Glauer, Stephan Günther, Frank Loebe, Oliver Kutz,  Till Mossakowski, Alan Ruttenberg, Barry Smith. 

\bibliography{./whatIsAnOntology}{}
\bibliographystyle{plain}

\end{document}